\documentclass{article}
\usepackage{flushend} 
\usepackage{balance}

\usepackage{microtype}
\usepackage{graphicx}
\usepackage{subfigure}
\usepackage{booktabs} 

\usepackage{hyperref}

\usepackage[accepted]{sty}
\usepackage{graphicx}
\usepackage{pythonhighlight}

\usepackage{times}
\usepackage{helvet}
\usepackage{courier}
\usepackage{subfigure} 
\usepackage{cancel}
\RequirePackage{amsthm,amsmath}
\usepackage{graphicx, amssymb, mathrsfs,amsfonts,url, bm}

\newcommand{\real}{\mathbb{R}}

\newcommand{\gap}{\,\,\,\,\,\,\,\,}

\newcommand{\blambda}{\boldsymbol\lambda}

\newcommand{\ba}{\bm{a}}
\newcommand{\bA}{\bm{A}}

\newcommand{\bE}{\bm{E}}

\newcommand{\bM}{\bm{M}}

\newcommand{\bw}{\bm{w}}
\newcommand{\bW}{\bm{W}}

\newcommand{\bz}{\bm{z}}
\newcommand{\bZ}{\bm{Z}}

\newcommand{\exponential}{\mathcal{E}}
\newcommand{\gammadist}{\mathcal{G}}
\newcommand{\inversegammadist}{\mathcal{G}^{-1}}
\newcommand{\rectifieddist}{\mathcal{RN}}
\newcommand{\normal}{\mathcal{N}}
\newcommand{\truncatednormal}{\mathcal{TN}}

\newcommand{\tnsng}{\mathcal{TNSNG}}
\newcommand{\rnsng}{\mathcal{RNSNG}}

\newcommand{\bmu}{\boldsymbol\mu}
\newcommand{\btau}{\boldsymbol\tau}

\usepackage[font=small,skip=2pt]{caption}
 
\newcommand{\titlethis}{Flexible and Hierarchical Prior for Bayesian Nonnegative Matrix Factorization}

\icmltitlerunning{\titlethis}

\begin{document}

\twocolumn[
\icmltitle{\titlethis}

\begin{icmlauthorlist}
	\icmlauthor{Jun Lu}{te}
	\icmlauthorsingle{\gap \gap Xuanyu Ye}
\end{icmlauthorlist}

\icmlaffiliation{te}{Correspondence to: Jun Lu $<$jun.lu.locky@gmail.com$>$. Copyright 2022 by the author(s)/owner(s). May 22nd, 2022}



\vskip 0.3in
]



\printAffiliationsAndNotice{}  

\begin{abstract}

In this paper, we introduce a probabilistic model for learning nonnegative
matrix factorization (NMF) that is commonly used for predicting missing values and finding hidden patterns in the data, in which the matrix factors are latent variables associated with each data dimension. The nonnegativity constraint for the latent factors is handled by choosing priors with support on the nonnegative subspace. Bayesian inference procedure based on Gibbs sampling is employed. We evaluate the model on several real-world datasets including MovieLens 100K and MovieLens 1M with different sizes and dimensions, and show that the proposed Bayesian NMF GRRN model leads to better predictions and avoids overfitting compared to existing Bayesian NMF approaches.

\end{abstract}
\section{Introduction}
Matrix factorization methods such as singular value decomposition, factor analysis, principal component analysis, and independent component analysis have been used extensively over the years to reveal hidden structures of matrices in many fields of science and engineering such as deep learning \citep{liu2015sparse}, recommendation systems \citep{comon2009tensor, lu2021numerical, lu2022matrix}, computer vision \citep{goel2020survey}, clustering and classification \citep{li2009non, wang2013non, lu2021survey}, collaborative filtering \citep{marlin2003modeling, lim2007variational, mnih2007probabilistic,raiko2007principal, chen2009collaborative} and machine learning in general \citep{lee1999learning}. 
Moreover, low-rank matrix approximation methods provide one of the simplest and most effective approaches to collaborative filtering for modeling user preferences. The idea behind such models is that preferences of a user are determined by a small number of unobserved factors \citep{salakhutdinov2008bayesian}.
In the meantime, nonnegative matrix factorization (NMF) models have been particularly popular since the constraint of nonnegativity makes the decompositional parts more interpretable \citep{lee1999learning} and have been studied extensively due to the popularity and prevalence of datasets like the Netflix challenge.

Methods for matrix factorization can be categorized as either non-probabilistic where the factorization is usually done by multiplicative updates \citep{comon2009tensor, lu2021numerical, lu2022matrix}, or probabilistic where the factorization is done by maximum-a-posteriori (MAP) or Bayesian inference \citep{mnih2007probabilistic, schmidt2009probabilistic, brouwer2017prior}. For the former, the solutions give a single point estimate that can lead to overfitting easily.
The MAP estimates resort to training models that amounts to maximizing the log-posterior over the model parameters. However the drawback is that inferring the posterior distribution over the factors is intractable \citep{hofmann1999probabilistic, marlin2003modeling, salakhutdinov2008bayesian}.
While Bayesian inference attacks this problem by finding a full distribution over the matrix spaces and reducing overfitting via different prior choices where the posterior is computed after observing the actual data.
This is important since the Bayesian approach has significant advantages in that it makes efficient use of the available data, allows prior information to be included into the model, avoids overfitting, allows a principled approach to model comparison, and allows missing data to be handled easily. 

In this light, we focus on Bayesian nonnegative factorizations of matrices.
The nonnegative factorization problem of observed matrix $\bA$ can be stated as $\bA=\bW\bZ+\bE$, where $\bA= [\ba_1, \ba_2, \ldots, \ba_N]\in \real^{M\times N}$ is approximately factorized into an $M\times K$ matrix $\bW\in \real_+^{M\times K}$ and a $K\times N$ matrix $\bZ\in \real_+^{K\times N}$ with nonnegative entries; the noise is captured by matrix $\bE\in \real^{M\times N}$. The dataset $\bA$ needs not to be complete such that the indices of observed entries can be represented by the mask matrix $\bM\in \{0,1\}^{M\times N}$. In the Netflix user preference context, this means that the $M\times N$ preference matrix of rating that $M$ users assign to $N$ movies is modeled by the product of an $M\times K$ user coefficient matrix $\bW$ and a $K\times N$ factor matrix $\bZ$ \citep{srebro2003weighted, salakhutdinov2008bayesian}. Training such models amounts to finding the best rank-$K$ approximation to the observed $M\times N$ target matrix $\bA$ under the given loss function.

Missing data is easily handled in the inference procedure
for the Bayesian NMF model by excluding the missing elements in the likelihood term.
Project data vectors $\ba_n$ to a smaller dimension $\bz_n \in \real^K$  with $K<M$,
such that the \textit{reconstruction error} measured by mean squared error (MSE or Frobenius norm) is minimized (assume $K$ is known):
\begin{equation}\label{equation:als-per-example-loss}
	\mathop{\min}_{\bW,\bZ} \,\, \sum_{n=1}^N \sum_{m=1}^{M} \left(a_{mn} - \bw_m^\top\bz_n\right)^2 \cdot m_{mn},
\end{equation}
where $\bW=[\bw_1^\top; \bw_2^\top; \ldots; \bw_M^\top]\in \real_+^{M\times K}$ and $\bZ=[\bz_1, \bz_2, \ldots, \bz_N] \in \real_+^{K\times N}$ contain $\bw_m$'s and $\bz_n$'s as \textbf{rows and columns} respectively, and $a_{mn}, m_{mn}$ are the $(m,n)$-th entries of data matrix $\bA$ and mask matrix $\bM$ respectively. The loss form in Eq.~\eqref{equation:als-per-example-loss} is known as the \textit{per-example loss}. And it can be equivalently written as
\begin{equation}\label{equation:loss_nmf_general}
\begin{aligned}
L(\bW,\bZ) &= \sum_{n=1}^N \sum_{m=1}^{M}\left(a_{mn} - \bw_m^\top\bz_n\right)^2 \cdot  m_{mn} \\
&= ||(\bW\bZ-\bA)\circledast \bM||^2 ,
\end{aligned}
\end{equation}
where $\circledast$ denotes the Hadamard product between matrices.

In this paper, we approach the nonnegative constraint by considering the NMF model as a latent factor model and we describe a fully specified graphical model for the problem and employ Bayesian learning methods to infer the latent factors. In this sense, explicit nonnegativity constraints are not required on the latent factors, since this is naturally taken care of by the appropriate choice of prior distribution, e.g., exponential density, half-normal density, truncated-normal density, or rectified-normal prior.

The main contribution of this paper is to propose a novel Bayesian NMF method which is flexible and has interpretable hyperpriors that shed light on the choice of hyperparameters in various applications.
We propose the hierarchical model called \textit{GRRN} NMF algorithm to both increase convergence performance and out-of-sample accuracy. While previous works propose somewhat algorithms that are also hierarchical (e.g., the GTTN in \citet{brouwer2017prior, schmidt2009probabilistic}), the methods still lack interpretability and are not flexible enough. 
The proposed GRRN model, a method for flexible and hierarchical Bayesian NMF, has simple conditional density forms with little extra computation. Meanwhile, the method is easy to implement. We show that our method can be successfully applied to the large, sparse, and very imbalanced Movie-User dataset, containing 100 million user/movie ratings.
We also show that the proposed GRRN model significantly increases the model’s predictive accuracy (held-our performance),
compared to the standard Bayesian NMF models and non-probabilistic NMF models.

\section{Related Work}

There are several probabilistic NMF methods. The one uses exponential priors to enforce nonnegativity (GEE model in \citet{schmidt2009bayesian}); the method uses truncated-normal priors to enforce nonnegativity (GTT model in \citet{brouwer2017prior}) and is further extended by a convenient joint hyperprior (GTTN model in \citet{schmidt2009probabilistic, brouwer2017prior}). 
GEE and GTT models are simply using exponential or truncated-normal priors to enforce nonnegativity in which case the models are sensitive to choices of hyperparameters. The GTTN model partly solves this problem while the posterior parameters are not easy to interpret such that the parameters of the hyperprior are not easy to be decided either (see Appendix~\ref{appendix:gibbs_other_NMF} for a detailed discussion on the inference). 
We propose a new hierarchical prior for Bayesian NMF models, which is designed to favor both flexibilities of parameter choices and interpretability of the parameters. 
The proposed GRRN model introduces an extra parameter that controls the behaviors of the prior densities on the latent variables. And this is key to increasing both the in-sample performance and out-of-sample accuracy.

\subsection{Probability Distributions}\label{section:probability_distribution}
We introduce all notations and probability distribution in this section.

$\normal(x|\mu, \tau^{-1}) =\sqrt{\frac{ \tau}{2\pi}}\exp\{-\frac{\tau}{2} (x-\mu)^2\}$ is a Gaussian distribution with mean $\mu$ and precision $\tau$ (variance $\sigma^2=\tau^{-1}$).

$\gammadist(x|\alpha, \beta)= \frac{\beta^\alpha}{\Gamma(\alpha)} x^{\alpha-1}\exp\{-\beta x\}u(x)$ is a Gamma distribution where $\Gamma(\cdot)$ is the gamma function and $u(x)$ is the unit step function that has a value of $1$ when $x\geq0$ and 0 otherwise.

$\inversegammadist(x|\alpha, \beta)= \frac{\beta^\alpha}{\Gamma(\alpha)} x^{-\alpha-1}\exp\{-\frac{\beta}{x}\}u(x)$ is an inverse-Gamma distribution.

$\exponential(x|\lambda) = \lambda\exp\{-\lambda x\}u(x)$ is an exponential distribution.

$\truncatednormal(x|\mu,\tau^{-1}) =\frac{\sqrt{\frac{\tau}{2\pi}} \exp\{-\frac{\tau}{2} (x-\mu)^2 \} } 
{1-\Phi(-\mu\sqrt{\tau})} u(x)$
is a truncated-normal (TN) with zero density below $x=0$ and renormalized to integrate to one. $\mu$ and $\tau$ are known as the ``parent mean" and ``parent precision". $\Phi(\cdot)$ is the cumulative distribution function of standard normal density $\normal(0,1)$.

$\rectifieddist(x| \mu, \tau^{-1}, \lambda) =\frac{1}{C} \cdot \normal(x|\mu, \tau^{-1})\cdot \exponential(x|\lambda) =\truncatednormal(x| \frac{\tau\mu-\lambda}{\tau} , \tau^{-1}) $ is known as the rectified-normal (RN) distribution with ``parent mean" $\mu$ and ``parent precision" $\tau$. The RN distribution is a more flexible distribution with zero density below $x=0$ than the TN distribution in the sense that it has an extra variable $\lambda$ to control the behavior of the density. $C$ is a constant given by 
$$
C(\mu, \tau, \lambda) = \lambda\left( 1-\Phi(-\frac{\tau\mu-\lambda}{\sqrt{\tau}}) \right) \exp\left\{-\mu\lambda+\frac{\lambda^2}{2\tau}\right\},
$$
which is a function with respect to $\{\mu, \tau, \lambda\}$.
Note in some texts, the TN distribution is termed as a RN distribution (e.g., \citet{schmidt2009probabilistic}). However, we here differentiate the two distributions where the reason will be clear in the sequel.

\section{Gaussian Rectified-Normal and Hierarchical Prior (GRRN) Model}

\subsection{Flexible and Hierarchical Model}

\begin{figure}[h]
\centering  
\subfigtopskip=2pt 
\subfigbottomskip=6pt 
\subfigcapskip=-15pt 
\includegraphics[width=0.35\textwidth]{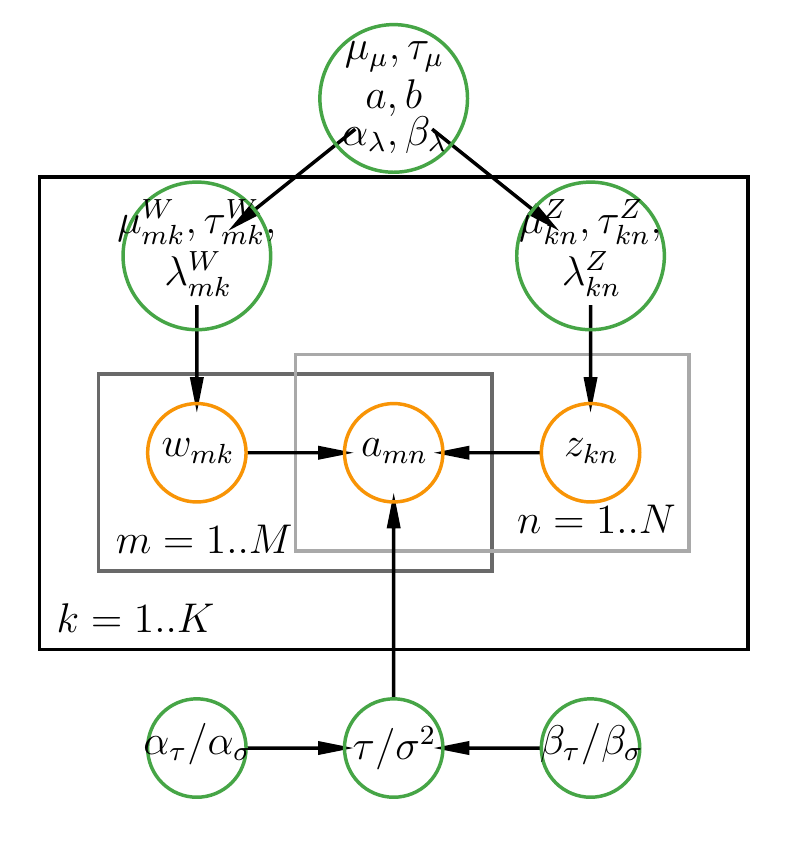}
\caption{Graphical representation of GRRN model. Orange circles represent observed and latent variables, green circles denote prior variables, and plates represent repeated variables. ``/" in the variable represents ``or", and comma ``," in the variable represents ``and".}
\label{fig:bmf_grrn}
\end{figure}

We view the data $\bA$ as being produced according to the probabilistic generative process shown in Figure~\ref{fig:bmf_grrn}. The observed data $(m,n)$-th entry $a_{mn}$ of matrix $\bA$ is modeled using a Gaussian likelihood function with variance $\sigma^2$ and mean given by the latent decomposition $\bw_m^\top\bz_n$ (Eq.~\eqref{equation:als-per-example-loss}),
\begin{equation}\label{equation:grrn_data_entry_likelihood}
p(a_{mn} | \bw_m^\top\bz_n, \sigma^2) = \normal(a_{mn}|\bw_m^\top\bz_n, \sigma^2).
\end{equation}
We choose a conjugate prior over the data variance, an inverse-Gamma distribution with shape $\alpha_\sigma$ and scale $\beta_\sigma$, 
\begin{equation}\label{equation:prior_grrn_gamma_on_variance}
	p(\sigma^2 | \alpha_\sigma, \beta_\sigma) = \inversegammadist(\sigma^2 | \alpha_\sigma, \beta_\sigma).
\end{equation}
While it can also be equivalently given a conjugate Gamma prior over the precision and we shall not repeat the details.

We treat the latent variables $w_{mk}$'s (and $z_{kn}$'s) as random variables. And we need prior densities over these latent variables to express beliefs for their values, e.g., nonnegativity in this context though there are many other constraints (semi-nonnegativity in \citet{ding2008convex}, discreteness in \citet{gopalan2014bayesian, gopalan2015scalable}).
Here we assume further that the latent variables $w_{mk}$'s are independently drawn from a rectified-normal prior
\begin{equation}\label{equation:rn_prior_grrn}
p(w_{mk} | \cdot ) = \rectifieddist(w_{mk} | \mu_{mk}^W, (\tau_{mk}^W)^{-1}, \lambda_{mk}^W).
\end{equation}
Similarly, the latent variables $z_{kn}$'s are also drawn from the same rectified-normal prior. This prior serves to enforce the nonnegativity constraint on the components $\bW, \bZ$, and is conjugate to the Gaussian likelihood.
In some scenarios, the two sets of latent variables can be drawn from two different rectified-normal priors, e.g., enforcing sparsity in $\bW$ while non-sparsity in $\bZ$. And we shall not consider this case as it is not the main interest of this paper. 
The posterior density is a truncated-normal distribution that is a special rectified-normal distribution.

To further favor flexibility, we choose a convenient joint hyperprior density over the parameters $\{\mu_{mk}^W, \tau_{mk}^W, \lambda_{mk}^W\}$ of RN prior in Eq.~\eqref{equation:rn_prior_grrn}, namely, the RN-scaled-normal-Gamma (RNSNG) prior,
\begin{equation}
\begin{aligned}
&\gap p(\mu_{mk}^W, \tau_{mk}^W, \lambda_{mk}^W |\cdot) \\
&= \rnsng(\mu_{mk}^W, \tau_{mk}^W, \lambda_{mk}^W| \mu_\mu, \tau_\mu, a, b, \alpha_\lambda, \beta_\lambda)\\
&=	C(\mu_{mk}^W, \tau_{mk}^W, \lambda_{mk}^W)
\cdot 
\normal(\mu_{mk}^W| \mu_\mu, (\tau_\mu)^{-1})\\
&\gap \cdot \gammadist(\tau_{mk}^W | a, b)
\cdot \gammadist(\lambda_{mk}^W | \alpha_\lambda, \beta_\lambda);
\end{aligned}
\end{equation}
This prior can decouple parameters $\mu_{mk}^W, \tau_{mk}^W, \lambda_{mk}^W$, and the posterior conditional densities of them are Gaussian, Gamma, and Gamma respectively due to this convenient scale.

\paragraph{Terminology} There are three types of choices we make that determine
the specific type of matrix decomposition model we use, namely, the likelihood function, the priors we place over the factor matrices $\bW$ and $\bZ$, and whether we use any further hierarchical priors. 
We will call the model by the density function in the order of the types of likelihood and priors. For example, if the likelihood function for the model is chosen to be a Gaussian density, and the two prior density functions are selected to be exponential density and Gaussian density functions respectively, then the model will be denoted as Gaussian Exponential-Gaussian (GEG) model. Sometimes, we will put a hyperprior over the parameters of the prior density functions, e.g., we put a Gamma prior over the Gaussian density, then it will further be termed as Gaussian Exponential-Gaussian Gamma (GEGA) model. In this sense, the proposed hierarchical model is named the \textit{GRRN} model.

\subsection{Gibbs Sampler}

\begin{algorithm}[tb] 
\caption{Gibbs sampler for GRRN. The procedure presented here may not be efficient but is explanatory. A more efficient one can be implemented in a vectorized manner. By default, uninformative priors are $\alpha_\sigma=\beta_\sigma=1$, $\mu_\mu =0$, $\tau_\mu=0.1, a=b=1$, $\alpha_\lambda=1, \beta_\lambda = \sqrt{\frac{m_0}{K}}$.} 
	\label{alg:grrn_gibbs_sampler}  
	\begin{algorithmic}[1] 
		\STATE {\bfseries Input:} Choose parameters $\alpha_\sigma, \beta_\sigma, \mu_\mu, \tau_\mu, a,  b, \alpha_\lambda, \beta_\lambda$;
\FOR{$t=1$ to $T$}
\FOR{$k=1$ to $K$} 
\FOR{$m=1$ to $M$}
		\STATE Sample $w_{mk}$ from Eq.~\eqref{equation:posterior_grrn_wmk};
		
		\STATE Sample $\mu_{mk}^W$ from Eq.~\eqref{equation:posterior_grrn_mu_tau1};
		
		\STATE Sample $\tau_{mk}^W$ from Eq.~\eqref{equation:posterior_grrn_tau_tau1};
		\STATE Sample $\lambda_{mk}^W$ from Eq.~\eqref{equation:posterior_grrn_lambda1};
		\ENDFOR
		\FOR{$n=1$ to $N$}
		\STATE Sample $z_{kn}$ from symmetry of Eq.~\eqref{equation:posterior_grrn_wmk};
		\STATE Sample $\mu_{kn}^Z$ from symmetry of Eq.~\eqref{equation:posterior_grrn_mu_tau1};
	\STATE Sample $\tau_{kn}^Z$ from symmetry of Eq.~\eqref{equation:posterior_grrn_tau_tau1};
	
	\STATE Sample $\lambda_{kn}^Z$ from symmetry of Eq.~\eqref{equation:posterior_grrn_lambda1};
	\ENDFOR
	\ENDFOR
	\STATE Sample $\sigma^2$ from $p(\sigma^2 | \bW,\bZ, \bA)$ in Eq.~\eqref{equation:posterior_grrn_sigma2}; 
\STATE Report loss in Eq.~\eqref{equation:loss_nmf_general}, stop if it converges.
\ENDFOR
\STATE Report mean loss in Eq.~\eqref{equation:loss_nmf_general} after burn-in iterations.
\end{algorithmic} 
\end{algorithm}

In this paper, we use Gibbs sampling since it tends to be very accurate at finding the true posterior. Other than this method, variational Bayesian inference can be an alternative way but we shall not go into the details. We shortly describe the posterior conditional density in this section. A detailed derivation can be found in Appendix~\ref{appendix:grrn_derivation}.
The conditional density for $\mu_{mk}^W$ is a truncated-normal (a special rectified-normal, see Section~\ref{section:probability_distribution}),
\begin{equation}\label{equation:posterior_grrn_mu_tau1}
\begin{aligned}
&\gap p(\mu_{mk}^W |\textcolor{black}{ \tau_{mk}^W, \lambda_{mk}^W}, \mu_\mu, \tau_\mu, a,  b,\alpha_\lambda, \beta_\lambda, w_{mk})\\
&\propto \rectifieddist(w_{mk} | \mu_{mk}^W, \frac{1}{\tau_{mk}^W}, \lambda_{mk}^W)
 C(\mu_{mk}^W, \tau_{mk}^W, \lambda_{mk}^W) \\
&\gap   \normal(\mu_{mk}^W| \mu_\mu, (\tau_\mu)^{-1})
 \gammadist(\tau_{mk}^W | a, b)
  \gammadist(\lambda_{mk}^W | \alpha_\lambda, \beta_\lambda)\\
&
\propto \normal(\mu_{mk}^W | \widetilde{m}, \widetilde{t}^{-1})\cdot u(\mu_{mk}^W)
\propto \normal(\mu_{mk}^W | \widetilde{m}, \widetilde{t}^{-1}),
\end{aligned}
\end{equation}
where 
$
\widetilde{t}=\tau_{mk}^W + \tau_\mu, 
\widetilde{m}=(\tau_{mk}^W w_{mk} +\tau_{\mu}\mu_\mu)/\widetilde{t}
$ are the posterior mean and precision respectively. The samples $w_{mk}$'s are nonnegative due to the rectification in the distribution (by exponential distribution inside the density). However, this mean parameter $\mu_{mk}^W$ is not limited to be nonnegative.

The conditional density for $\tau_{mk}^W$ is a Gamma distribution,
\begin{equation}\label{equation:posterior_grrn_tau_tau1}
\begin{aligned}
&\gap p(\tau_{mk}^W |\textcolor{black}{ \mu_{mk}^W, \lambda_{mk}^W}, \mu_\mu, \tau_\mu, a,  b,\alpha_\lambda, \beta_\lambda, w_{mk})\\
&\propto \rectifieddist(w_{mk} | \mu_{mk}^W, \frac{1}{\tau_{mk}^W}, \lambda_{mk}^W)
 C(\mu_{mk}^W, \tau_{mk}^W, \lambda_{mk}^W) \\
&\gap   \normal(\mu_{mk}^W| \mu_\mu, (\tau_\mu)^{-1})
 \gammadist(\tau_{mk}^W | a, b)
  \gammadist(\lambda_{mk}^W | \alpha_\lambda, \beta_\lambda)\\
&\propto \gammadist(\tau_{mk}^W | \widetilde{a}, \widetilde{b}),
\end{aligned}
\end{equation}
where $\widetilde{a} = a+\frac{1}{2}, \widetilde{b}= b+ \frac{(w_{mk}-\mu_{mk}^W)^2}{2}$ are the posterior shape and scale parameters.

Furthermore, the conditional density for $\lambda_{mk}^W$ is also a Gamma distribution,
\begin{equation}\label{equation:posterior_grrn_lambda1}
\begin{aligned}
&\gap p(\lambda_{mk}^W |\textcolor{black}{ \mu_{mk}^W,\tau_{mk}^W} , \mu_\mu, \tau_\mu, a,  b,\alpha_\lambda, \beta_\lambda, w_{mk})\\
&\propto \rectifieddist(w_{mk} | \mu_{mk}^W, \frac{1}{\tau_{mk}^W}, \lambda_{mk}^W)
 C(\mu_{mk}^W, \tau_{mk}^W, \lambda_{mk}^W) \\
&\gap   \normal(\mu_{mk}^W| \mu_\mu, (\tau_\mu)^{-1})
 \gammadist(\tau_{mk}^W | a, b)
  \gammadist(\lambda_{mk}^W | \alpha_\lambda, \beta_\lambda)\\
& \propto \gammadist(\lambda_{mk}^W | \widetilde{\alpha_\lambda}, \widetilde{\beta_\lambda}),
\end{aligned}
\end{equation}
where 
$
\widetilde{\alpha_\lambda}= \alpha_\lambda+1, 
\widetilde{\beta_\lambda}= \beta_\lambda + w_{mk}
$.
The importance for this hierarchical prior is revealed that,
from this conditional density, the prior parameter $\alpha_\lambda$ can be interpreted as the number of prior observations, and $\beta_\lambda$ as the prior value of $w_{mk}$. On the one hand, an uninformative choice for $\alpha_\lambda$ is $\alpha_\lambda=1$. On the other hand, if one prefers a sparse decomposition with larger regularization on the model, $\beta_\lambda$ can be chosen as a small value, e.g., $\beta_\lambda=0.01$. Or a large value, e.g., $\beta_\lambda=100$, can be applied since we are in the NMF context; a large value in $\bW$ will enforce the counterparts in $\bZ$ to have small values. While, an \textit{uninformative choice} for $\beta_\lambda$ is as follows. Suppose the mean of matrix $\bA$ is $m_0$, then $\beta_\lambda$ can be set as $\beta_\lambda=\sqrt{\frac{m_0}{K}}$ where the $K$ is the latent dimension such that each prior entry $a_{mn}=\bw_m^\top\bz_n$ is equal to $m_0$.  After developing this hierarchical prior, we realize its similarity with the GTTN model (first introduced in a tensor decomposition context \citep{schmidt2009probabilistic}, and further discussed in \citet{brouwer2017prior}). However, the parameters in conditional densities of GTTN model lack interpretation and flexibility so that there is no guidelines for parameter tuning when the performance is poor. The proposed GRRN model, on the other hand, can work well generally when we select the uninformative prior $\beta_\lambda=\sqrt{\frac{m_0}{K}}$; moreover, one can even set $\beta_\lambda=20 \cdot \sqrt{\frac{m_0}{K}}$ or $0.1 \cdot \sqrt{\frac{m_0}{K}}$ if one prefers a larger regularization as mentioned above.

\begin{figure*}[h]
	\centering  
	\subfigtopskip=2pt 
\subfigbottomskip=2pt 
\subfigcapskip=2pt 
\subfigure[Convergence on the \textbf{MovieLens 100K} dataset with increasing latent dimension $K$.]{\includegraphics[width=1\textwidth]{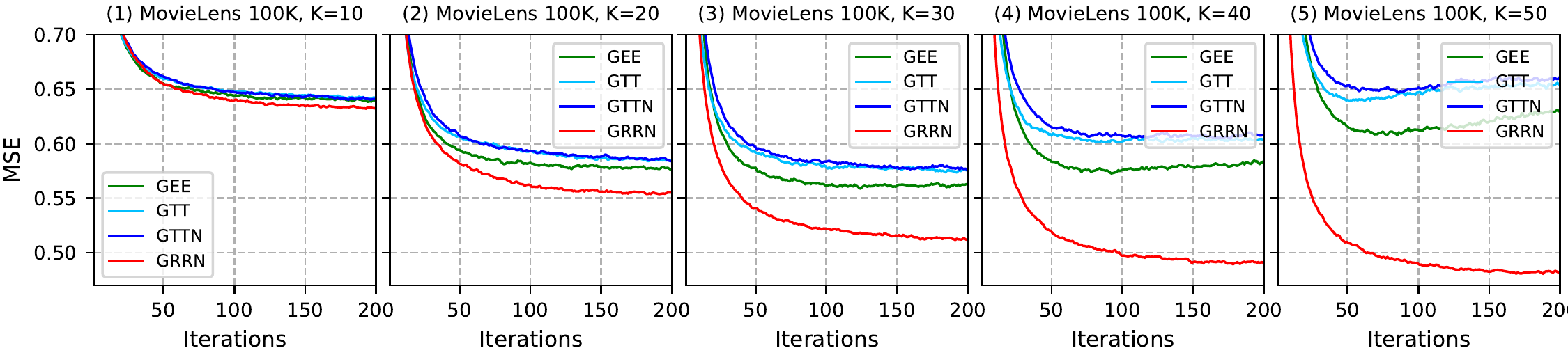} \label{fig:convergences_gdsc_20}}
\subfigure[Convergence on the \textbf{MovieLens 1M} dataset with increasing latent dimension $K$.]{\includegraphics[width=1\textwidth]{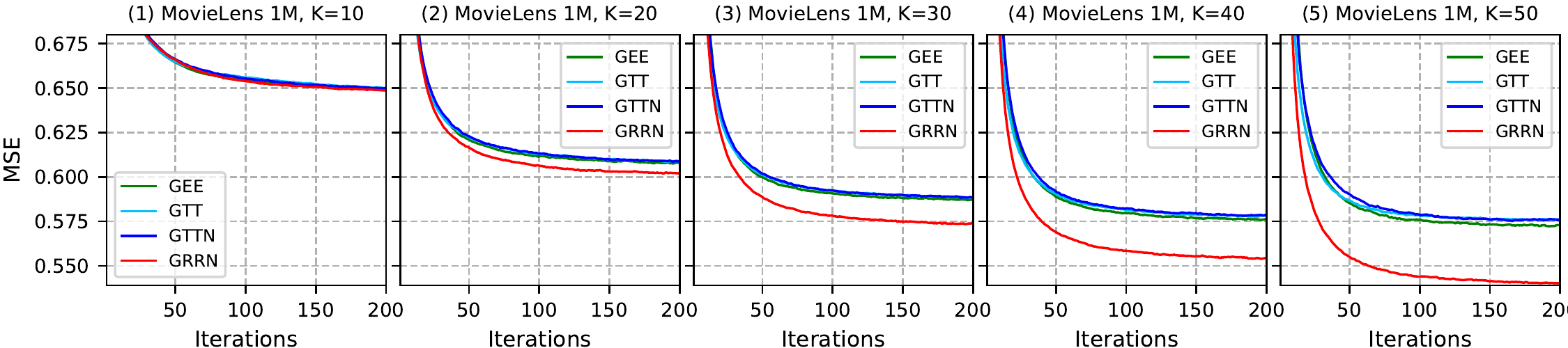} \label{fig:convergences_movielens100k_20}}
\caption{Convergence of the models on the 
MovieLens 100K (upper) and the MovieLens 1M (lower) datasets, measuring the training data fit (mean squared error). When increasing latent dimension $K$, the proposed GRRN continues to increase the performance; while other models start to decrease on the MovieLens 100K dataset or stop increasing on the MovieLens 1M dataset.}
\label{fig:convergences_gdsc_movielens100k}
\end{figure*}

The conditional density of $w_{mk}$ is a truncated-normal density. Denote all elements of $\bW$ except $w_{mk}$ as $\bW_{-mk}$, we then have
\begin{equation}\label{equation:posterior_grrn_wmk}
\begin{aligned}
&\gap p(w_{mk}| \sigma^2, \bW_{-mk}, \bZ, \mu_{mk}^W, \tau_{mk}^W,\lambda_{mk}^W)\\
&\propto p(\bA|\bW, \bZ, \sigma^2) \times p(w_{mk} | \mu_{mk}, (\tau_{mk})^{-1}, \lambda_{mk})\\
&\propto \prod_{i,j=1}^{M,N} \normal(a_{ij}| \bw_i^\top\bz_j, \sigma^2) 
 \truncatednormal(\frac{\tau_{mk}^W\mu_{mk}^W - \lambda_{mk}^W}{\tau_{mk}^W}, \frac{1}{\tau_{mk}} )\\
& 
\propto \truncatednormal(w_{mk}| \widetilde{\mu_{mk}} , \widetilde{\sigma_{mk}^2}),
\end{aligned}
\end{equation}
where $\widetilde{\sigma_{mk}^2} = \frac{\sigma^2}{ \sum_{j=1}^{N} z_{kj}^2 + \tau_{mk}^W \cdot \sigma^2 }$ is the posterior ``parent variance" of the normal distribution with posterior ``parent mean" 
$$
\widetilde{\mu_{mk}} = 
\left(
\frac{1}{\sigma^2} \sum_{j=1}^{N}z_{kj}(a_{mj}- \sum_{i\neq k}^{K}w_{mk}z_{ij}) + \tau_{mk}^W \mu^\prime 
\right)\cdot \widetilde{\sigma_{mk}^2}
$$
with $\mu^\prime = \frac{\tau_{mk}^W\mu_{mk}^W - \lambda_{mk}^W}{\tau_{mk}^W}$ being the ``parent mean" of the truncated-normal density \footnote{To simplify matters, we assume that there are no missing ratings firstly. 
Note here if the data contains missing entries, the summation in $\widetilde{\mu_{mk}}$ contains only the observed ones. Thus the algorithm can be used to predict the missing entries.}.

Due to symmetry, the conditional expression for $z_{kn}$, $\mu_{kn}^Z$, $\tau_{kn}^Z$, and $\lambda_{kn}^Z$ can be easily derived similarly; and we shall not go into the details.

Finally, the conditional density of $\sigma^2$ is an inverse-Gamma distribution by conjugacy,
\begin{equation}\label{equation:posterior_grrn_sigma2}
\begin{aligned}
&\gap p(\sigma^2 | \bW, \bZ, \bA)
= \inversegammadist(\sigma^2 | \widetilde{\alpha_\sigma}, \widetilde{\beta_\sigma}),
\end{aligned}
\end{equation}
where $\widetilde{\alpha_\sigma} = \frac{MN}{2}+\alpha_\sigma$, 
$\widetilde{\beta_\sigma}=\frac{1}{2} \sum_{i,j=1}^{M,N}(a_{ij}-\bw_i^\top\bz_j)^2+\beta_\sigma$.
The full procedure is formulated in Algorithm~\ref{alg:grrn_gibbs_sampler}.

\subsection{Computational Complexity}
The adopted Gibbs sampling method for GRRN model has complexity $\mathcal{O}(MNK^2)$ where the most costs come from the update on the conditional density of $w_{mk}$ and $z_{kn}$. In the meantime, all the methods we use to compare the results (GEE, GTT, GTTN) have complexity $\mathcal{O}(MNK^2)$. Compared to the GTTN model, the proposed GRRN model only has an extra cost on the update of $\lambda_{mk}^W$ which does not amount to the bottleneck of the algorithm.

\begin{table}[H]
	\begin{tabular}{llll}
		\hline
		Dataset        & Rows & Columns & Fraction obs. \\ \hline
		MovieLens 100K & 943  & 1473    & 0.072         \\ 
		MovieLens 1M &6040 &3503& 0.047 \\
		\hline
	\end{tabular}
	\caption{Dataset description. 99,723 and 999,917 observed entries for MovieLens 100K and MovieLens 1M datasets respectively (user vectors or movie vectors with less than 3 observed entries are cleaned). MovieLens 100K is relatively a small dataset and the MovieLens 1M tends to be large; while both of them are sparse.}
	\label{table:datadescription}
\end{table}
\begin{figure}[H]
	\centering  
	\subfigtopskip=2pt 
	\subfigbottomskip=2pt 
	\subfigcapskip=-5pt 
	\subfigure{\includegraphics[width=0.231\textwidth]{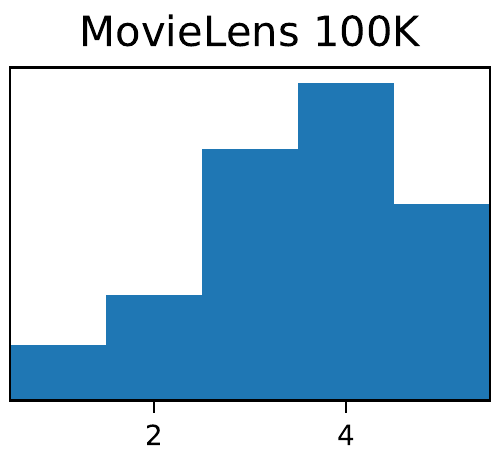} \label{fig:data_movielen100k}}
	\subfigure{\includegraphics[width=0.231\textwidth]{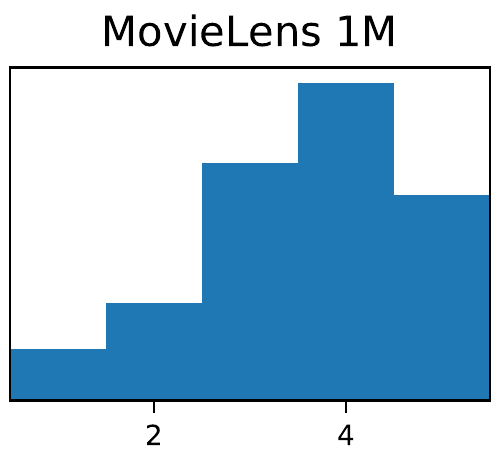} \label{fig:data_movielen1m}}
	\caption{Data distribution of MovieLens 100K and MovieLens 1M datasets. 
		The MovieLens 1M dataset has a larger fraction of users who give a rate of 5 and a smaller fraction for rates of 3.}
	\label{fig:datasets_nmf}
\end{figure}
\section{Experiments}\label{section:ader_experiments}
To evaluate the strategy and demonstrate the main advantages of the proposed GRRN method, 
we conduct experiments with different analysis tasks; and different data sets including the MovieLens 100K
and the MovieLens 1M from movie ratings for different users \citep{harper2015movielens}. 
The datasets have a range from one to five stars with around 100,000 and 1,000,000 ratings respectively and we want to predict the missing entries for users so that we can recommend the movies they like (user vectors or movie vectors with less than 3 observed entries are cleaned).
A summary of the two datasets can be seen in Table~\ref{table:datadescription} and their distributions are shown in Figure~\ref{fig:datasets_nmf}.
The MovieLens 1M dataset has a larger fraction of users who give a rate of 5 
and a smaller fraction for rates of 3.
We can see that the MovieLens 100K is relatively a small dataset and the MovieLens 1M tends to be large; while both of them are sparse. On the other hand, the MovieLens 1M dataset not only has a larger number in the users, but also has an increased dimension (the number of movies) making it a harder task to evaluate.

In all scenarios, same parameter initialization is adopted when conducting different tasks. 
We compare the results in terms of convergence speed and generalization. In a wide range of scenarios across various models, GRRN improves convergence rates, and leads to out-of-sample performances that are as good or better than existing Bayesian NMF models. 

\subsection{Hyperparameters}
We follow the default hyperparameter setups in \citet{brouwer2017prior}. We use $\{\lambda_{mk}^W\}=\{\lambda_{kn}^Z\}=0.1$ (GEE); $\{\mu_{mk}^Z\}=\{\mu_{kn}^Z\}=0, \{\tau_{mk}^Z\}=\{\tau_{kn}^Z\}=0.1$ (GTT); uninformative $\alpha_\sigma=\beta_\sigma=1$ (Gaussian likelihood in GEE, GTT, GTTN, GRRN); 
$\mu_\mu =0$, $\tau_\mu=0.1, a=b=1$ (hyperprior in GTTN, GRRN); $\alpha_\lambda=1, \beta_\lambda = \sqrt{\frac{m_0}{K}}$ (hyperprior in GRRN). These are very weak prior choices and the models are not sensitive to them \citep{brouwer2017prior}.
As long as the hyperparameters are set, the observed or unobserved variables are initialized from random draws as this initialization procedure provides a better initial guess of the right patterns in the matrices.
In all experiments, we run the Gibbs sampler 500 iterations with a burn-in of 400 iterations as the convergence analysis shows the algorithm can converge in less than 200 iterations.

\subsection{Convergence Analysis}
Firstly we compare the convergence in terms of iterations on the MovieLens 100K and MovieLens 1M datasets. We run each model with $K=10, 20, 30, 40, 50$, and the loss is measured by mean squared error (MSE).
Figure~\ref{fig:convergences_gdsc_movielens100k} shows the average convergence results of ten repeats. 
On the MovieLens 1M dataset, all the methods converge to better performance with smaller MSE when increasing latent dimension $K$; while the performance of GRRN is better than the other models. Moreover, we observe that the convergence results of GTT and GTTN models are rather close since they share similar hidden structures though GTTN is a hierarchical model.
On the other hand, when conducting on the MovieLen 100K dataset and increasing the feature dimension $K$, the GRRN model continues to converge to better performance with MSE continuing to decrease. However, GEE, GTT, and GTTN models first converge to a better performance and then start to diverge with larger MSE observed or stop improving at all when increasing latent dimension $K$. From this perspective, GRRN is a better choice for data reduction compared to other Bayesian NMF models.

\subsection{Noise Sensitivity}
\begin{figure}[h]
\centering  
\subfigtopskip=2pt 
\subfigbottomskip=9pt 
\subfigcapskip=-5pt 
\includegraphics[width=0.35\textwidth]{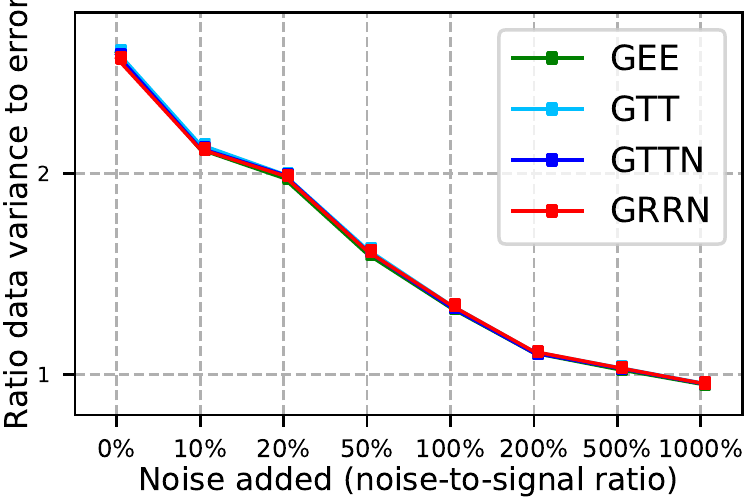}
\caption{Ratio of the variance of data to the MSE of the predictions. The higher the better.}
\label{fig:noise_graph_movielens100k}
\end{figure}

We further measure the noise sensitivity of different models with predictive performance when the datasets are noisy. To see this, we add different levels of Gaussian noise to the data. We add levels of $\{0\%, 10\%,$ $20\%,$ $50\%, 100\%, 200\%, 500\%, 1000\%\}$ noise-to-signal ratio noise (which is the ratio of the variance of the added Gaussian noise to the variance of the data). The results for the MovieLens 100K with $K=50$ are shown in Figure~\ref{fig:noise_graph_movielens100k}. We observe that the proposed GRRN model performs similarly to other NMF models. Similar results can be found on the MovieLens 1M dataset and other $K$ values and we shall not repeat the details.

\begin{table}[]
	\begin{tabular}{lllll}
		\hline
		$K$\textbackslash{}Models & GEE & GTT   & GTTN & GRRN   \\ \hline
		$K$=20   &  1.18   &  1.06    &   1.07  &   \textbf{ 1.02 } \\
		$K$=30   &  1.43   &  1.18    &   1.20  &   \textbf{ 1.00 } \\
		$K$=40   &  1.86   &  1.42    &   1.45  &   \textbf{ 0.98 } \\
		$K$=50   &  2.63   &  1.84    &   1.89  &   \textbf{ 0.97 } \\
		\hline
		$K$=20   &  3.47   &  1.46    &   1.57  &   \textbf{ 1.10 } \\
		$K$=30   &  6.86   &  2.27    &   2.52  &   \textbf{ 1.05 } \\
		$K$=40   &  17056.27   &  4.07    &   4.79  &   \textbf{ 1.04 } \\
		$K$=50   &  236750.39   &  2650.21    &   5452.18  &   \textbf{ 1.05 } \\
		\hline
\end{tabular}
\caption{Mean squared error measure when 97\% (upper table) and 98\% (lower table) of data is unobserved for MovieLens 100K dataset. The performance of the proposed GRRN model is only a little worse when increasing the fraction of unobserved from 97\% to 98\%.
Similar situations can be observed in the MovieLens 1M experiment.}
\label{table:movielens100k_special_sparsity_case}
\end{table}

\begin{figure*}[h]
\centering  
\subfigtopskip=2pt 
\subfigbottomskip=2pt 
\subfigcapskip=-2pt 
\subfigure[Predictive results on the \textbf{MovieLens 100K} dataset with increasing fraction of unobserved data and increaing latent dimension $K$.]{\includegraphics[width=1\textwidth]{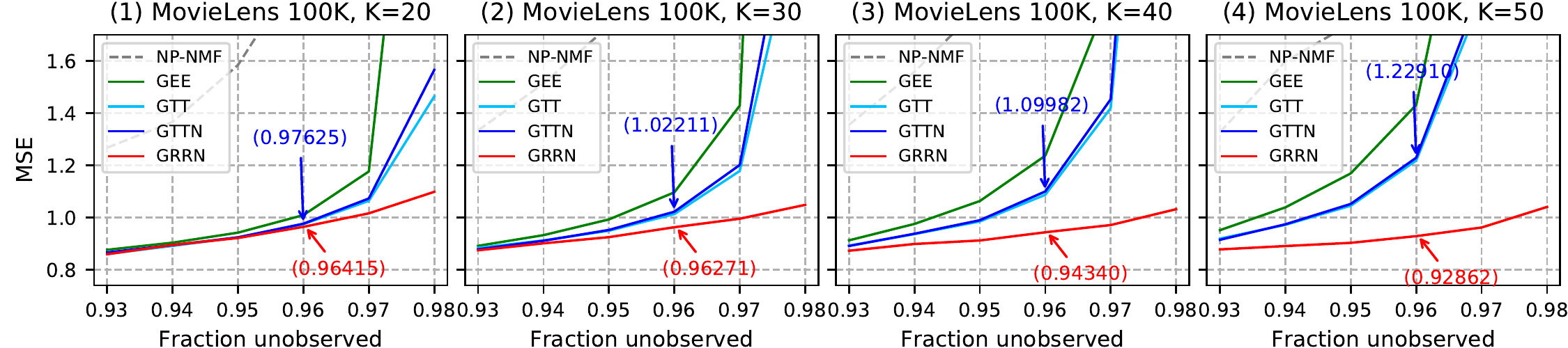} \label{fig:sparsity_movielens_100k_variousK}}
\subfigure[Predictive results on the \textbf{MovieLens 1M} dataset  with increasing fraction of unobserved data and increasing latent dimension $K$.]{\includegraphics[width=1\textwidth]{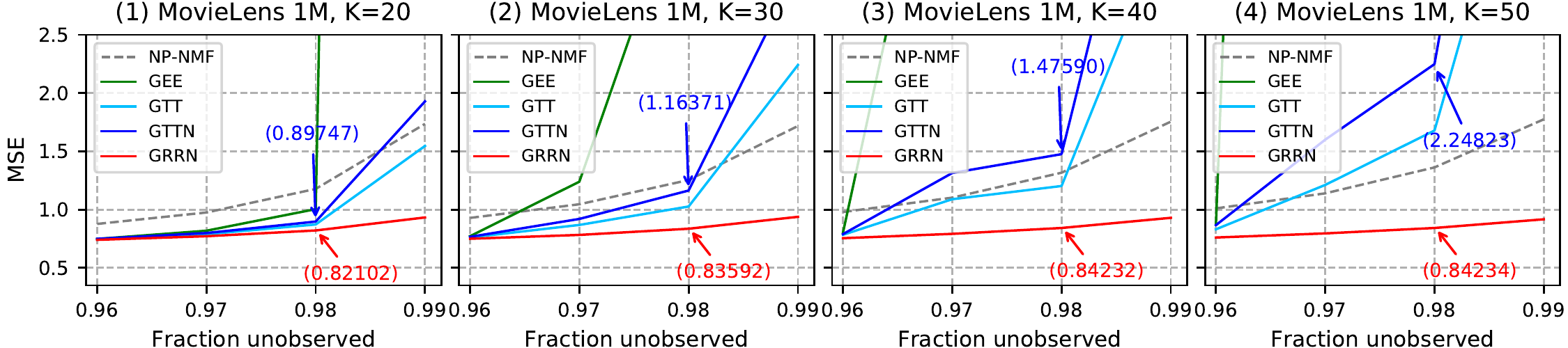} \label{fig:sparsity_movielens_1M_variousK}}
\caption{Predictive results on the MovieLens 100K (upper) and MovieLens 1M (lower) datasets with the least fractions of unobserved data being 0.928 and 0.953 respectively (Table~\ref{table:datadescription}).
We measure the predictive performance (mean squared error) on a held-out dataset for different fractions of unobserved data. The blue and red arrows compare the MSEs of GTTN and GRRN models when the fractions of unobserved data are 0.96 and 0.98 respectively. }
	\label{fig:sparsity_movielen100_1M}
\end{figure*}

\subsection{Predictive Analysis}
The training performance of the GRRN model steadily improves as the model complexity grows. Inspired by this result, 
we measure the predictive performance when the sparsity of the data increases to see whether the models overfit or not. For different fractions of unobserved data, we randomly split the data based on that fraction, train the model on the observed data, and measure the performance on the held-out test data. Again, we increase $K$ from $K=20$ to $K=30, 40, 50$ for all models. The average MSE of ten repeats is given in Figure~\ref{fig:sparsity_movielen100_1M}. We observe that when $K=20$ and the fraction of unobserved data is relatively a small value 
(e.g., fraction unobserved = 0.93 in Figure~\ref{fig:sparsity_movielens_100k_variousK} and fraction unobserved = 0.96 in Figure~\ref{fig:sparsity_movielens_1M_variousK}), 
all the models perform similarly (GRRN is only slightly better). However, when the fraction of unobserved data increases or the latent dimension $K$ increases, GRRN performs much better than the other models.

Table~\ref{table:movielens100k_special_sparsity_case} shows MSE predictions of different models when the fraction of unobserved data is $97\%$ and $98\%$. We observe that the performance of the proposed GRRN model is only a little worse when increasing the fraction of unobserved from 97\% to 98\% showing the proposed GRRN model is more robust with less overfitting. While for other competitive models, the performances become extremely worse in this scenario.
From Figure~\ref{fig:convergences_gdsc_movielens100k}, we see that GEE can converge to a better in-sample performance generally; this leads to a worse out-of-sample performance as shown in Figure~\ref{fig:sparsity_movielen100_1M} (compared to GTT and GTTN). However, the proposed GRRN has both better in-sample and out-of-sample performances from this experiment making it a more robust choice in predicting missing entries. Similar situations can be observed in the MovieLens 1M case.

We also add a popular non-probabilistic NMF (NP-NMF) model to see the predictive results \citep{lee2000algorithms}. Empirical results (grey lines in Figure~\ref{fig:sparsity_movielen100_1M}) show that the NP-NMF can overfit easily compared to Bayesian NMF approaches even when the fraction of unobserved data is relatively small and latent dimension $K$ is small though the issue is less severer in the MovieLens 1M dataset.

\section{Conclusion}
The aim of this paper is to solve the issue of hyperparameter choice in the Bayesian NMF context for collaborative filtering, factor analysis, and many other problems. We propose a simple and computationally efficient algorithm that requires little extra computation and is easy to implement for nonnegative matrix factorization. Overall, we show that the proposed GRRN model is a versatile algorithm that has better convergence results and out-of-sample performances on both small and large datasets. 
GRRN is able to avoid problems of overfitting which is common in the standard non-probabilistic NMF model and other Bayesian NMF models.


\bibliography{bib}
\bibliographystyle{sty}
\balance

\onecolumn
\appendix

\section{Derivation of Gibbs Sampler for GRRN Model}\label{appendix:grrn_derivation}
In this section, we give the derivation of Gibbs sampler for the proposed GRRN model. As shown in the main paper, the observed data $(m,n)$-th entry $a_{mn}$ of matrix $\bA$ is modeled using a Gaussian likelihood with variance $\sigma^2$ and mean given by the latent decomposition $\bw_m^\top\bz_n$ (Eq.~\eqref{equation:als-per-example-loss}),
\begin{equation}\label{equation:grrn_data_entry_likelihood_app}
	p(a_{mn} | \bw_m^\top\bz_n, \sigma^2) = \normal(a_{mn}|\bw_m^\top\bz_n, \sigma^2).
\end{equation}
We choose a conjugate prior on the data variance, an inverse-Gamma distribution with shape $\alpha_\sigma$ and scale $\beta_\sigma$, 
\begin{equation}\label{equation:prior_grrn_gamma_on_variance_app}
	p(\sigma^2 | \alpha_\sigma, \beta_\sigma) = \inversegammadist(\sigma^2 | \alpha_\sigma, \beta_\sigma).
\end{equation}
We assume further that the latent variable $w_{mk}$'s are drawn from a rectified-normal prior,
\begin{equation}
p(w_{mk} | \cdot ) = \rectifieddist(w_{mk} | \mu_{mk}^W, (\tau_{mk}^W)^{-1}, \lambda_{mk}^W)=
\truncatednormal\left(x|\frac{\tau_{mk}^W\mu_{mk}-\lambda_{mk}^W}{\tau_{mk}^W}, (\tau_{mk}^W)^{-1}\right).
\end{equation}
Similarly, the latent variable $z_{kn}$'s are also drawn from the same rectified-normal prior. This prior serves to enforce the nonnegativity constraint on the components $\bW, \bZ$, and is conjugate to the Gaussian likelihood.
Moreover, we place a RN-scaled-normal-Gamma hyperprior over the parameters of the rectified-normal prior,
$$
\begin{aligned}
\mu_{mk}^W, \tau_{mk}^W, \lambda_{mk}^W | \mu_\mu, \tau_\mu, a,  b, \alpha_\lambda, \beta_\lambda
&\sim \rnsng(\mu_{mk}^W, \tau_{mk}^W, \lambda_{mk}^W | \mu_\mu, (\tau_\mu)^{-1}, a,  b, \alpha_\lambda, \beta_\lambda)\\
& \propto   C(\mu_{mk}^W, \tau_{mk}^W, \lambda_{mk}^W)  \cdot 
\normal(\mu_{mk}^W| \mu_\mu, (\tau_\mu)^{-1})
\cdot \gammadist(\tau_{mk}^W | a, b)
\cdot \gammadist(\lambda_{mk}^W | \alpha_\lambda, \beta_\lambda);\\
\mu_{kn}^Z, \tau_{kn}^Z, \lambda_{kn}^Z | \mu_\mu, \tau_\mu, a,  b, \alpha_\lambda, \beta_\lambda
&\sim \rnsng(\mu_{kn}^Z, \tau_{kn}^Z, \lambda_{kn}^Z | \mu_\mu, (\tau_\mu)^{-1}, a,  b, \alpha_\lambda, \beta_\lambda).
\end{aligned}
$$
Following from the graphical representation of the GRRN model in Figure~\ref{fig:bmf_grrn}, 
the conditional density of $\mu_{mk}^W$ can be derived,
\begin{equation}\label{equation:posterior_grrn_mu_tau1222_app}
	\begin{aligned}
		&\gap p(\mu_{mk}^W |\textcolor{black}{ \tau_{mk}^W, \lambda_{mk}^W}, \mu_\mu, \tau_\mu, a,  b,\alpha_\lambda, \beta_\lambda, w_{mk},\cancel{\bmu_{-mk}^W}, \cancel{\btau_{-mk}^W})\\
		&\propto \rectifieddist(w_{mk} | \mu_{mk}^W, (\tau_{mk}^W)^{-1}, \lambda_{mk}^W)
\times C(\mu_{mk}^W, \tau_{mk}^W, \lambda_{mk}^W) 
		\cdot  \normal(\mu_{mk}^W| \mu_\mu, (\tau_\mu)^{-1})
		\cdot \gammadist(\tau_{mk}^W | a, b)
		\cdot  \gammadist(\lambda_{mk}^W | \alpha_\lambda, \beta_\lambda)\\
		&= 
		\normal(w_{mk}| \mu_{mk}^W, (\tau_{mk}^W)^{-1})\cdot 
		\cancel{\exponential(w_{mk}| \lambda_{mk}^W)}
		\cdot  \normal(\mu_{mk}^W| \mu_\mu, (\tau_\mu)^{-1})
		\cdot \cancel{\gammadist(\tau_{mk}^W | a, b)}
		\cdot  \cancel{\gammadist(\lambda_{mk}^W | \alpha_\lambda, \beta_\lambda)}\\
		&\propto \normal(w_{mk}| \mu_{mk}^W, (\tau_{mk}^W)^{-1})\normal(\mu_{mk}^W| \mu_\mu, (\tau_\mu)^{-1})
		\propto \normal(\mu_{mk}^W | \widetilde{m}, \widetilde{t}^{-1}),
	\end{aligned}
\end{equation}
where 
$
\widetilde{t}=\tau_{mk}^W + \tau_\mu, 
\widetilde{m}=(\tau_{mk}^W w_{mk} +\tau_{\mu}\mu_\mu)/\widetilde{t}.
$
The conditional density of $\tau_{mk}^W$ is,
\begin{equation}\label{equation:posterior_grrn_tau_tau1222_app}
	\begin{aligned}
		&\gap p(\tau_{mk}^W |\textcolor{black}{ \mu_{mk}^W, \lambda_{mk}^W}, \mu_\mu, \tau_\mu, a,  b,\alpha_\lambda, \beta_\lambda, w_{mk},\cancel{\bmu_{-mk}^W}, \cancel{\btau_{-mk}^W})\\
		&\propto \rectifieddist(w_{mk} | \mu_{mk}^W, (\tau_{mk}^W)^{-1}, \lambda_{mk}^W)\times C(\mu_{mk}^W, \tau_{mk}^W, \lambda_{mk}^W) 
		\cdot  \normal(\mu_{mk}^W| \mu_\mu, (\tau_\mu)^{-1})
		\cdot \gammadist(\tau_{mk}^W | a, b)
		\cdot  \gammadist(\lambda_{mk}^W | \alpha_\lambda, \beta_\lambda)\\
		&= 
		\normal(w_{mk}| \mu_{mk}^W, (\tau_{mk}^W)^{-1})\cdot 
		\cancel{\exponential(w_{mk}| \lambda_{mk}^W)}
		\cdot  \cancel{\normal(\mu_{mk}^W| \mu_\mu, (\tau_\mu)^{-1})}
		\cdot {\gammadist(\tau_{mk}^W | a, b)}
		\cdot  \cancel{\gammadist(\lambda_{mk}^W | \alpha_\lambda, \beta_\lambda)}\\
&\propto \normal(w_{mk}| \mu_{mk}^W, (\tau_{mk}^W)^{-1})\gammadist(\tau_{mk}^W | a, b)\\
		&\propto (\tau_{mk}^W)^{a+\frac{1}{2}-1} \exp\left\{-\left( b+ \frac{(w_{mk}-\mu_{mk}^W)^2}{2} \right) \tau_{mk}^W \right\}  
		\propto \gammadist(\tau_{mk}^W | \widetilde{a}, \widetilde{b}),
	\end{aligned}
\end{equation}
where $\widetilde{a} = a+\frac{1}{2}, \widetilde{b}= b+ \frac{(w_{mk}-\mu_{mk}^W)^2}{2}$.
And the conditional density of $\lambda_{mk}^W$ is,
\begin{equation}\label{equation:posterior_grrn_lambda122_app}
	\begin{aligned}
		&\gap p(\lambda_{mk}^W |\textcolor{black}{ \mu_{mk}^W,\tau_{mk}^W} , \mu_\mu, \tau_\mu, a,  b,\alpha_\lambda, \beta_\lambda, w_{mk},\cancel{\bmu_{-mk}^W}, \cancel{\btau_{-mk}^W})\\
		&\propto \rectifieddist(w_{mk} | \mu_{mk}^W, (\tau_{mk}^W)^{-1}, \lambda_{mk}^W)\times C(\mu_{mk}^W, \tau_{mk}^W, \lambda_{mk}^W) 
		\cdot  \normal(\mu_{mk}^W| \mu_\mu, (\tau_\mu)^{-1})
		\cdot \gammadist(\tau_{mk}^W | a, b)
		\cdot  \gammadist(\lambda_{mk}^W | \alpha_\lambda, \beta_\lambda)\\
		&= 
		\cancel{\normal(w_{mk}| \mu_{mk}^W, (\tau_{mk}^W)^{-1})}\cdot 
		{\exponential(w_{mk}| \lambda_{mk}^W)}
		\cdot  \cancel{\normal(\mu_{mk}^W| \mu_\mu, (\tau_\mu)^{-1})}
		\cdot \cancel{\gammadist(\tau_{mk}^W | a, b)}
		\cdot  {\gammadist(\lambda_{mk}^W | \alpha_\lambda, \beta_\lambda)}\\
		&\propto \exponential(w_{mk}| \lambda_{mk}^W)\gammadist(\lambda_{mk}^W | \alpha_\lambda, \beta_\lambda) \propto \gammadist(\lambda_{mk}^W | \widetilde{\alpha_\lambda}, \widetilde{\beta_\lambda}),
	\end{aligned}
\end{equation}
where 
$
\widetilde{\alpha_\lambda}= \alpha_\lambda+1,  
\widetilde{\beta_\lambda}= \beta_\lambda + w_{mk}.
$
And the prior parameter $\alpha_\lambda$ can be interpreted as the number of prior observations, and $\beta_\lambda$ as the prior value of $w_{mk}$. If one prefers a sparse decomposition, $\beta_\lambda$ can be chosen with a small value, e.g., $0.1$.
Or a large value, e.g., $\beta_\lambda=100$, can be applied since we are in the NMF context; a large value in $\bW$ will enforce the counterparts in $\bZ$ to have small values.
Finally, the conditional density of $w_{mk}$ is,
\begin{equation}
\begin{aligned}
&\gap p(w_{mk}| \sigma^2, \bW_{-mk}, \bZ, \bmu^W, \btau^W,\blambda^W, \cancel{\bmu^Z},\cancel{\btau^Z} , \cancel{\blambda^Z})
\propto p(\bA|\bW, \bZ, \sigma^2) \times p(w_{mk} | \mu_{mk}, \tau_{mk}, \lambda_{mk})\\
&\propto \prod_{i,j=1}^{M,N} \normal(a_{ij}| \bw_i^\top\bz_j, \sigma^2) 
\times \rectifieddist(\mu_{mk}, (\tau_{mk})^{-1}, \lambda_{mk})
\propto \prod_{i,j=1}^{M,N} \normal(a_{ij}| \bw_i^\top\bz_j, \sigma^2) 
\times \truncatednormal\left( \underbrace{\frac{\tau_{mk}^W\mu_{mk}^W - \lambda_{mk}^W}{\tau_{mk}^W}}_{\mu^\prime}, (\tau_{mk})^{-1} \right)\\
&\propto 
\exp\left\{ -\left(\frac{\sum_{j=1}^{N}z_{kj}^2 }{2\sigma^2} + \frac{\tau_{mk}^W}{2}\right)w_{mk}^2 + w_{mk} \left(
\frac{1}{\sigma^2} \sum_{j=1}^{N}z_{kj}(a_{mj}- \sum_{i\neq k}^{K}w_{mk}z_{ij}) + \tau_{mk}^W \mu^\prime 
\right)  \right\} u(w_{mk})\\
& \propto \normal(w_{mk}| \widetilde{\mu_{mk}} , \widetilde{\sigma_{mk}^2})u(w_{mk}) 
= \truncatednormal(w_{mk}| \widetilde{\mu_{mk}} , \widetilde{\sigma_{mk}^2}),
\end{aligned}
\end{equation}
where $\widetilde{\sigma_{mk}^2} = \frac{\sigma^2}{ \sum_{j=1}^{N} z_{kj}^2 + \tau_{mk}^W \cdot \sigma^2 }$ is the posterior ``parent variance" of the normal distribution with posterior ``parent mean" 
$$
\widetilde{\mu_{mk}} = 
\left(
\frac{1}{\sigma^2} \sum_{j=1}^{N}z_{kj}(a_{mj}- \sum_{i\neq k}^{K}w_{mk}z_{ij}) + \tau_{mk}^W \mu^\prime 
\right)\cdot \widetilde{\sigma_{mk}^2}.
$$
with $\mu^\prime = \frac{\tau_{mk}^W\mu_{mk}^W - \lambda_{mk}^W}{\tau_{mk}^W}$ being the ``parent mean" of the truncated-normal density. The full procedure is already formulated in Algorithm~\ref{alg:grrn_gibbs_sampler}.

\section{Gibbs Sampling Algorithms of Other NMF Models}\label{appendix:gibbs_other_NMF}
In this section, we give the posteriors for competitive Bayesian NMF models that we will compare in the paper, namely, the GEE, GTT, and GTTN models.
\subsection{Gaussian Likelihood with Exponential Priors (GEE) Model}
\begin{figure}[h]
	\centering  
	\vspace{-0.35cm} 
	\subfigtopskip=2pt 
	\subfigbottomskip=2pt 
	\subfigcapskip=-5pt 
	\subfigure[GEE.]{\label{fig:bmf_gee}
	\includegraphics[width=0.321\linewidth]{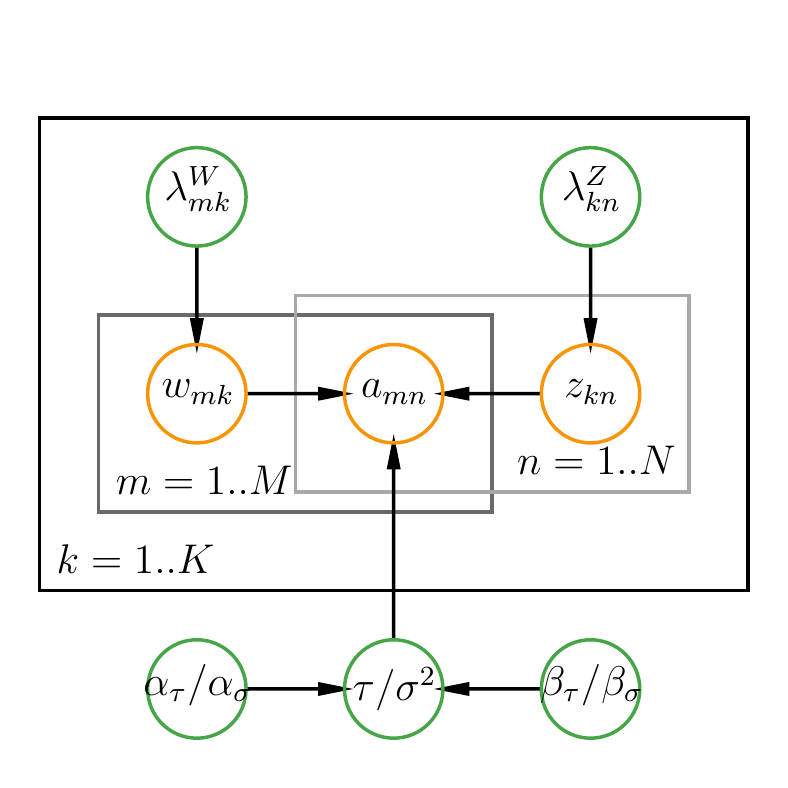}}
	\subfigure[GTT.]{\label{fig:bmf_gtt}
		\includegraphics[width=0.321\linewidth]{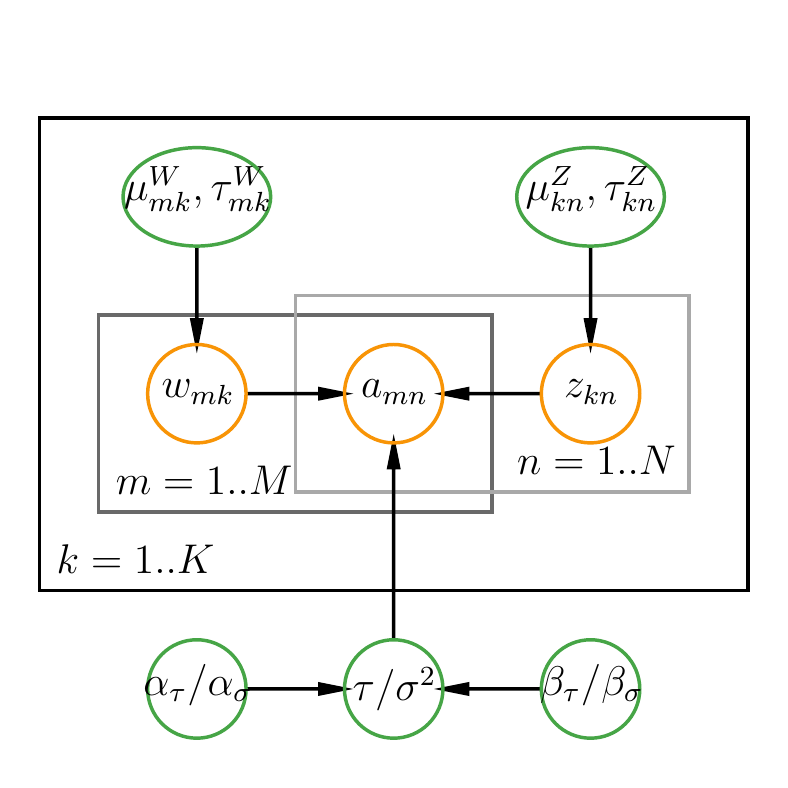}}
	\subfigure[GTTN.]{\label{fig:bmf_gttn}
		\includegraphics[width=0.321\linewidth]{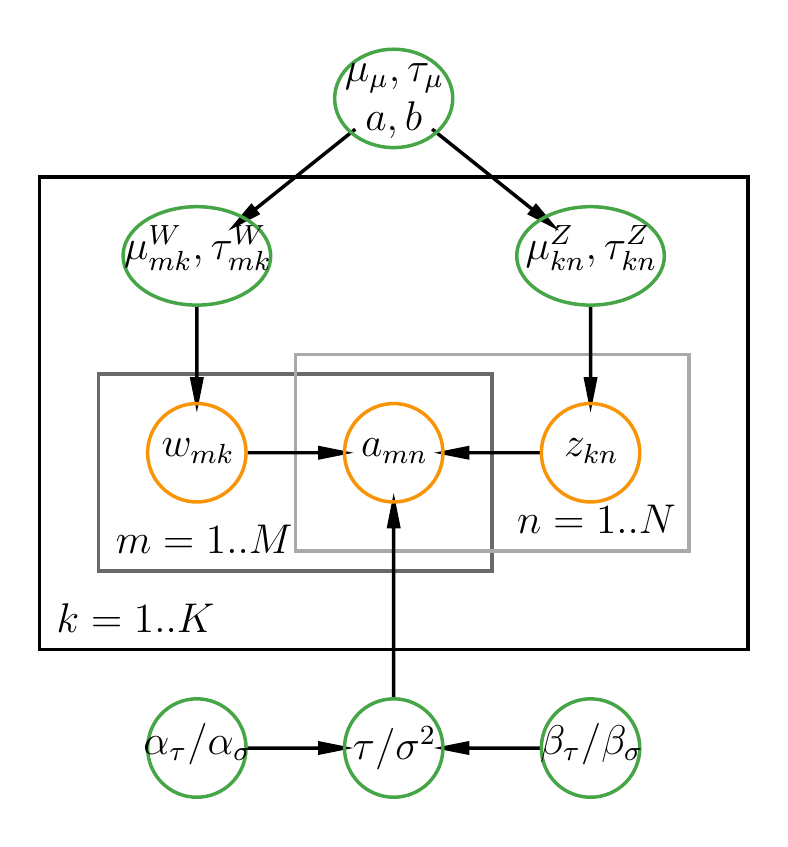}}
\caption{Graphical model representation of GEE, GTT, and GTTN models. Orange circles represent observed and latent variables, green circles denote prior variables, and plates represent repeated variables. ``/" in the variable represents ``or", and comma ``," in the variable represents ``and".}
	\label{fig:bmf_gtt_gttn}
\end{figure}
The Gaussian Exponential-Exponential (GEE) model is perhaps the simplest one for Bayesian NMF \citep{schmidt2009bayesian} where each entry $a_{mn}$ of matrix $\bA$ is again modeled using a Gaussian likelihood with variance $\sigma^2$ and mean given by the latent decomposition $\bw_m^\top\bz_n$ (Eq.~\eqref{equation:grrn_data_entry_likelihood}, this will be default for GTT and GTTN models as well). The graphical representation is shown in Figure~\ref{fig:bmf_gee}.
The model places independent exponential priors over the entries of $\bW, \bZ$, 
$$
w_{mk}\sim \exponential(w_{mk}| \lambda_{mk}^W), \qquad 
z_{kn}\sim \exponential(z_{kn}| \lambda_{kn}^W).
$$
The product of a Gaussian and an exponential distribution leads to a truncated-normal posterior,
$$
p(w_{mk} | \sigma^2, \bW_{-mk}, \bZ, \lambda_{mk}^W, \bA) 
= \truncatednormal(w_{mk} | \widetilde{\mu_{mk}}, \widetilde{\sigma_{mk}^{2}}),
$$
where $\widetilde{\sigma_{mk}^{2}}= \frac{\sigma^2}{\sum_{j=1}^{N} z_{kj}^2}$ is the posterior ``parent variance" of the normal distribution with mean $\widetilde{\mu_{mk}}$, that is,
$$
\widetilde{\mu_{mk}} = \left( -\lambda_{mk}^W+ \frac{1}{\sigma^2} \sum_{j=1}^{N} z_{kj}\bigg( a_{mj} - \sum_{i\neq k}^{K}w_{mi}z_{ij}\bigg)  \right)\cdot \widetilde{\sigma_{mk}^{2}}
$$
is the posterior ``parent mean" of the normal distribution, and $\truncatednormal(x | \mu, \sigma^2)$ is the \textit{truncated normal density} with ``parent mean" $\mu$ and ``parent variance" $\sigma^2$.
Or after rearrangement, the posterior density of $w_{mk}$ can be equivalently described by 
$$
p(w_{mk} | \sigma^2, \bW_{-mk}, \bZ, \lambda, \bA) 
=  \rectifieddist(w_{mk} |   \widehat{\mu_{mk}}, \widehat{\sigma_{mk}^{2}}, \lambda_{mk}^W ),
$$
where $\widehat{\sigma^2_{mk}}=\widetilde{\sigma_{mk}^{2}}= \frac{\sigma^2}{\sum_{j=1}^{N} z_{kj}^2}$ is the posterior ``parent variance" of the normal distribution with ``parent mean" $\widehat{\mu_{mk}}$, 
$$
\widehat{\mu_{mk}} =  \frac{1}{\sum_{j=1}^{N} z_{kj}^2} \cdot \sum_{j=1}^{N} z_{kj}\bigg( a_{mj} - \sum_{i\neq k}^{K}w_{mi}z_{ij}\bigg) .
$$
Due to symmetry, a similar expression for $z_{kn}$ can be easily derived. And the condition density of $\sigma^2$ is the same as that in the GRRN model (Eq.~\eqref{equation:posterior_grrn_sigma2}). 

\subsection{Gaussian Likelihood with Truncated-Normal Priors (GTT) Model}
The Gaussian Truncated-Normal (GTT) model is discussed in \citet{brouwer2017prior} where
entry $a_{mn}$ of matrix $\bA$ is again modeled using a Gaussian likelihood with variance $\sigma^2$ and mean given by the latent decomposition $\bw_m^\top\bz_n$ (Eq.~\eqref{equation:grrn_data_entry_likelihood})
 and truncated-normal priors are used over factor matrices.
The graphical representation of the GTT model is shown in Figure~\ref{fig:bmf_gtt}.
We use the truncated-normal distribution as the priors for entries of $\bW, \bZ$, 
$$
w_{mk} \sim \truncatednormal(w_{mk} | \mu_{mk}^{W}, (\tau_{mk}^W)^{-1} ), \qquad  
z_{kn} \sim \truncatednormal(z_{kn} | \mu_{kn}^Z, (\tau_{kn}^Z)^{-1} ). 
$$
The product of a Gaussian and a truncated-normal distribution leads to a truncated-normal density as well, 
$$
p(w_{mk} | \sigma^2, \bW_{-mk}, \bZ,  \mu_{mk}^W, (\tau_{mk}^W)^{-1}, \bA) 
= \truncatednormal(w_{mk} | \widetilde{\mu_{mk}}, \widetilde{\sigma_{mk}^{2}}),
$$
where $\widetilde{\sigma_{mk}^{2}}= \frac{\sigma^2}{\sum_{j=1}^{N} z_{kj}^2 + \tau_{mk}^W\cdot \sigma^2}$ is the posterior ``parent variance" of the normal distribution with mean $\widetilde{\mu_{mk}}$, that is,
$$
\widetilde{\mu_{mk}} = \left( \frac{1}{\sigma^2} \sum_{j=1}^{N} z_{kj}\bigg( a_{mj} - \sum_{i\neq k}^{K}w_{mi}z_{ij}  \bigg)  + \textcolor{black}{\tau_{mk}^W \mu_{mk}^W}\right)\cdot \widetilde{\sigma_{mk}^{2}}
$$
is the posterior ``parent mean" of the normal distribution.
Again, due to symmetry, a similar expression for $z_{kn}$ can be easily derived. And the condition density of $\sigma^2$ is the same as that in the GRRN model (Eq.~\eqref{equation:posterior_grrn_sigma2}).

\subsection{Gaussian Likelihood with Truncated-Normal and Hierarchical Priors (GTTN) Model}
Similar to the GRRN model, we can further prior over the parameters of the truncated-normal distributions (Figure~\ref{fig:bmf_gttn}) based on the GTT model, hence the name GTTN model. The hierarchical prior is proposed in \citet{schmidt2009probabilistic} and further discussed in \citet{brouwer2017prior}. A convenient prior called \textit{TN-scaled-normal-Gamma} distribution is used (or a \textit{TN-scaled-normal-inverse-Gamma} prior for mean and variance parameters): 
$$
\begin{aligned}
	\mu_{mk}^W, \tau_{mk}^W | \mu_\mu, \tau_\mu, a,  b
	&\sim \tnsng(\mu_{mk}^W, \tau_{mk}^W | \mu_\mu, (\tau_\mu)^{-1}, a,  b)\\
	&\propto  \frac{1}{\sqrt{\tau_{mk}^W}} \left(1 - \Phi\big(-\mu_{mk}^W\sqrt{\tau_{mk}^W} \big) \right) \cdot 
	\normal(\mu_{mk}^W| \mu_\mu, (\tau_\mu)^{-1})\cdot 
	\gammadist(\tau_{mk}^W | a, b);\\
	\mu_{kn}^Z, \tau_{kn}^Z | \mu_\mu, \tau_\mu, a,  b
	&\sim \tnsng(\mu_{kn}^Z, \tau_{kn}^Z | \mu_\mu, (\tau_\mu)^{-1}, a,  b).
\end{aligned}
$$
The conditional density of $\mu_{mk}^W$ is 
\begin{equation}\label{equation:posterior_gttn_mu_tau1}
\begin{aligned}
&\gap p(\mu_{mk}^W |\textcolor{black}{ \tau_{mk}^W}, \mu_\mu, \tau_\mu, a,  b, w_{mk},\cancel{\bmu_{-mk}^W}, \cancel{\btau_{-mk}^W})\\
&\propto \truncatednormal(w_{mk} | \mu_{mk}^W, (\tau_{mk}^W)^{-1} )
\cdot  \frac{1}{\sqrt{\tau_{mk}^W}} \left(1 - \Phi\big(-\mu_{mk}^W\sqrt{\tau_{mk}^W} \big) \right)  
\normal(\mu_{mk}^W| \mu_\mu, (\tau_\mu)^{-1}) 
\gammadist(\tau_{mk}^W | a, b)\\
&
\propto \normal(\mu_{mk}^W | \widetilde{m}, \widetilde{t}^{-1}),
\end{aligned}
\end{equation}
where 
$
\widetilde{t}=\tau_{mk}^W + \tau_\mu, 
\widetilde{m}=(\tau_{mk}^W w_{mk} +\tau_{\mu}\mu_\mu)/\widetilde{t}.
$
And the conditional density of $\tau_{mk}^W$ is 
\begin{equation}\label{equation:posterior_gttn_mu_tau2}
\begin{aligned}
&\gap p(\tau_{mk}^W |\textcolor{black}{\mu_{mk}^W }, \mu_\mu, \tau_\mu, a,  b, w_{mk},\cancel{\bmu_{-mk}^W}, \cancel{\btau_{-mk}^W})\\
&\propto \truncatednormal(w_{mk} | \mu_{mk}^W, (\tau_{mk}^W)^{-1} )
\cdot  \frac{1}{\sqrt{\tau_{mk}^W}} \left(1 - \Phi\big(-\mu_{mk}^W\sqrt{\tau_{mk}^W} \big) \right)  
\normal(\mu_{mk}^W| \mu_\mu, (\tau_\mu)^{-1}) 
\gammadist(\tau_{mk}^W | a, b)\\
&
\propto \gammadist(\tau_{mk}^W | \widetilde{a}, \widetilde{b}),
\end{aligned}
\end{equation}
where $\widetilde{a} = a+\frac{1}{2}, \widetilde{b}= b+ \frac{(w_{mk}-\mu_{mk}^W)^2}{2}$.
And again due to symmetry, the expressions for $\mu_{kn}^Z$ and $\tau_{kn}^Z$ can be easily derived similarly.

\end{document}